\documentclass[times,final]{elsarticle}
\usepackage{times}
\usepackage{amsmath,amsfonts,amssymb}
\usepackage{graphicx}
\usepackage{booktabs,multirow}
\usepackage{latexsym}
\usepackage[small]{caption}
\usepackage{subcaption}
\usepackage{graphicx,booktabs,colortbl,multirow}
\usepackage[table,x11names,dvipsnames,table]{xcolor}
\usepackage{listings}
\newtheorem{definition}{Definition}

\newtheorem{example}{Example}

\usepackage{hyperref}

\usepackage{color} %% Highlight
\usepackage{float}
\usepackage{tikz}

\begin{document}

\makeatletter
\def\ps@pprintTitle{%
  \let\@oddhead\@empty
  \let\@evenhead\@empty
  \let\@oddfoot\@empty
  \let\@evenfoot\@oddfoot
}
\makeatother

\begin{frontmatter}
\title{A Topological Approach for Semi-Supervised Learning}

\author{A. Inés$^{1}$\corref{mycorrespondingauthor}}\ead{adrian.ines@unirioja.es}
\author{C. Domínguez$^{1}$, J. Heras$^{1}$, G. Mata$^{1}$ and J. Rubio$^{1}$}
% \small{$^{1}$Department of Mathematics and Computer Science, Unversity of La Rioja, Spain}\\
% \small{$^{2}$Department of Biomedical Engineering, School of Biomedical Engineering}\\\small{ $\&$ Imaging Sciences, King’s Collegue London. UK.}\\
% }

\address{$^{1}$University of La Rioja, Department of Mathematics and Computer Science, Spain
}

\cortext[mycorrespondingauthor]{Corresponding author}
%
% \titlerunning{Semi-Supervised Compact Networks in the Medical Context}
% If the paper title is too long for the running head, you can set
% an abbreviated paper title here
%
%\author{First Author\inst{1}\orcidID{0000-1111-2222-3333} \and
%Second Author\inst{2,3}\orcidID{1111-2222-3333-4444} \and
%Third Author\inst{3}\orcidID{2222--3333-4444-5555}}
%
%\authorrunning{F. Author et al.}
% First names are abbreviated in the running head.
% If there are more than two authors, 'et al.' is used.
%
%\institute{Princeton University, Princeton NJ 08544, USA \and
%Springer Heidelberg, Tiergartenstr. 17, 69121 Heidelberg, Germany
%\email{lncs@springer.com}\\
%\url{http://www.springer.com/gp/computer-science/lncs} \and
%ABC Institute, Rupert-Karls-University Heidelberg, Heidelberg, Germany\\
%\email{\{abc,lncs\}@uni-heidelberg.de}}
%             % typeset the header of the contribution
%
\begin{abstract}
Nowadays, Machine Learning and Deep Learning methods have become the state-of-the-art approach to solve data classification tasks. In order to use those methods, it is necessary to acquire and label a considerable amount of data; however, this is not straightforward in some fields, since data annotation is time consuming and might require expert knowledge. This challenge can be tackled by means of semi-supervised learning methods that take advantage of both labelled and unlabelled data. In this work, we present new semi-supervised learning methods based on techniques from Topological Data Analysis (TDA), a field that is gaining importance for analysing large amounts of data with high variety and dimensionality. In particular, we have created two semi-supervised learning methods following two different topological approaches. In the former, we have used a homological approach that consists in studying the persistence diagrams associated with the data using the Bottleneck and Wasserstein distances. In the latter, we have taken into account the connectivity of the data. In addition, we have carried out a thorough analysis of the developed methods using 3 synthetic datasets, 5 structured datasets, and 2 datasets of images. The results show that the semi-supervised methods developed in this work outperform both the results obtained with models trained with only manually labelled data, and those obtained with classical semi-supervised learning methods, reaching improvements of up to a 16\%.

% \textbf{Keywords: } Topological Data Analysis, Semi-supervised learning, Bottleneck distance, Wasserstein distance, Data connectivity; 
\end{abstract}


\end{frontmatter}
\section{Introduction}
Machine Learning and Deep Learning techniques have become the state-of-the-art approach to solve classification problems in a wide variety of fields such as biology~\cite{Affonso17}, security~\cite{security}, or medicine~\cite{breastCancer}. One of the main problems of these techniques is the great amount of data that they need to work properly~\cite{dataDem}. This may not seem a problem due to the large amount of data that is being generated in a daily basis. However, data acquisition is not easy in some fields due to, for example, a limited budget to obtain samples, the need to perform an invasive medical procedure or destructive processes. In addition, in supervised learning, one of the  main approaches in machine learning, the data has to be annotated, and it is well-known that this might be a problem because it is a very time-consuming task that might require expert knowledge~\cite{anotationDem}.

% The Machine and Deep Learning community has already tackle this challenge using techniques like data augmentation~\cite{dataaugm} or semi-supervised learning~\cite{Laine17,Berthelot19}. 

Semi-supervised learning methods~\cite{Berthelot19,Laine17} have received growing attention in recent years to tackle this challenge. These methods provide a mean of using unlabelled data to improve models' performance when we have access to a large corpus of data that is difficult to annotate. Traditional semi-supervised learning algorithms, such as  Label Spreading~\cite{labelSpread} or Label Propagation~\cite{labelProp}, focus on the distance between the data points to annotate unlabelled data points; that is, on the metric and density characteristics of the data in a dataset. However, topological characteristics of the data are not used, and this is the approach proposed in this paper.

Topological Data Analysis (from now on, TDA) has arisen as a field to extract topological and geometrical information from data, to reveal dynamical organisation of the brain~\cite{brainTDA}, to recognising atmospheric river patterns in large climate datasets~\cite{atmosphericTDA}, or to examine spreading processes on networks~\cite{networksTDA}. An important result of TDA is the Manifold Hypothesis~\cite{fefferman2016testing}, that states that high dimensional data tends to lie in low dimensional manifolds, and that has inspired our definition of semi-supervised learning methods for binary classification tasks. Intuitively, our methods are based on the following idea. Given two sets of data points $A$ and $B$, we can define two manifolds associated with each set, ${\cal M}_A$ and ${\cal M}_B$ respectively. Now, given an unlabelled data point $x$ that belongs to either $A$ or $B$; if $x$ belongs to $A$, analogously for $B$, then the manifold associated with $A\cup \{x\}$ will be more similar ${\cal M}_A$ than if we compare the manifold associated with $B\cup \{x\}$ and ${\cal M}_B$. The rest of the paper is devoted to introduce this idea formally; namely, the contributions of this work are the following:

\begin{itemize}
\item We present several semi-supervised methods based on TDA notions.
\item We conduct a thorough analysis for our methods and compare their performance with classical semi-supervised learning methods. To this aim, we have employed a benchmark composed of 10 different datasets (3 synthetic datasets, 5 structured datasets, and 2 datasets of images).
\item We introduce a library that allows users to employ our methods.
\end{itemize}

The rest of this paper is organised as follows. In the next section, we provide the necessary background to understand the rest of the paper. Subsequently, we present our semi-supervised learning methods in Section~\ref{sec:methods}, the datasets used for evaluating our methods in Section~\ref{sec:eval}, and the results of our experiments in Section~\ref{sec:results}. The paper ends with a section of conclusions and further work. This work has an associated project webpage where the interested reader can consult all the code and examples presented in this paper: \url{https://github.com/adines/TTASSL}.

\section{Background}
This work can be framed in the context of both Topological Data Analysis and semi-supervised learning. In this section, we introduce the necessary notions of these fields to understand the methods proposed in this work. For a more detailed introduction to TDA see~\cite{zomorodian2012topological}, and for semi-supervised learning see~\cite{zhu2009introduction}.

\subsection{Topological Data Analysis}
Topological Data Analysis is a field that aims to extract information about data based on its topology. The most widely used tool in TDA is persistent homology~\cite{computationalTop,persist,computingPersistHomology}, which allows us to measure certain features of a space, such as its connectivity, holes or voids. All these features are based on the concept of simplicial complex.

\begin{definition}[Simplicial complex]
Let $V$ be a finite nonempty set whose elements are called vertices. A \textit{simplicial complex} on $V$ is a collection $K$ of nonempty subsets of $V$ subject to two requirements:
\begin{itemize}
    \item for each vertex $v$ in $V$, the singleton $\{v\}$ is in $K$, and
    \item if $\tau$ is in $K$ and $\sigma \subset \tau$ then $\sigma$ must also be in $K$.
\end{itemize}
Given two simplicial complexes $K_1$ and $K_2$, if $K_1 \subset K_2$ then $K_1$ is called subcomplex of $K_2$.
\end{definition}

\begin{example}
Let us consider the set $V=\{1,2,3,4\}$, the simplicial complex represented in Figure~\ref{fig:example1} is $K=\{1,2,3,4,12,13,23,14,24,123\}$. A subcomplex of $K$ is, for instance, $K=\{1,2,3,12,13,23,123\}$

\begin{figure}[ht]
\centering
	\includegraphics[width=0.5\textwidth]{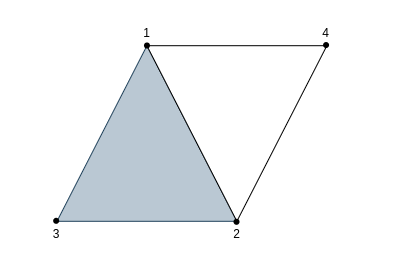}
    \caption{Example of a simplicial complex.}
    \label{fig:example1}
\end{figure}

\end{example}

In this work, we want to study the topological properties of a dataset; so, we have to build a simplicial complex from a dataset. To this aim, each point of the dataset is represented as a point in an $n$-dimensional space, where $n$ is the number of features of the point. Then, we can construct the Vietoris-Rips complex as follows.

\begin{definition}[Vietoris-Rips complex]
Let $(M,d)$ be a finite metric space. For every $\epsilon>0$, the \textit{Vietoris-Rips complex} $VR_\epsilon$ is defined as follows:
\[
VR_\epsilon(M)=\{\sigma\subseteq M \mid \forall u,v \in \sigma : d(u,v)\leq \epsilon\}
\]
\end{definition}

We can notice that in the previous definition we do not have a single simplicial complex, but rather we have a set of simplicial complexes that depend on $\epsilon$, a value called the radius. Such a sequence of simplicial complexes is a called a filtration. 

\begin{definition}[Filtration]
Let $K$ be a simplicial complex. A \textit{filtration} of $K$ (of length $n$) is a nested sequence of subcomplexes of the form
\[
K_1\subset K_2\subset\cdots\subset K_{n-1}\subset K_n=K 
\]
\end{definition}

\begin{example}
Let us consider the points $x_1=(0,0)$, $x_2=(3,0)$ and $x_3=(2,2)$ in the Euclidian space. The Vietoris-Rips complex for three different values of $\epsilon$ ($\epsilon=0.5$, $\epsilon=2.5$, and $\epsilon=2.9$) can be seen in Figure~\ref{fig:example2}.

\begin{figure}[ht]
\centering
	\includegraphics[width=\textwidth]{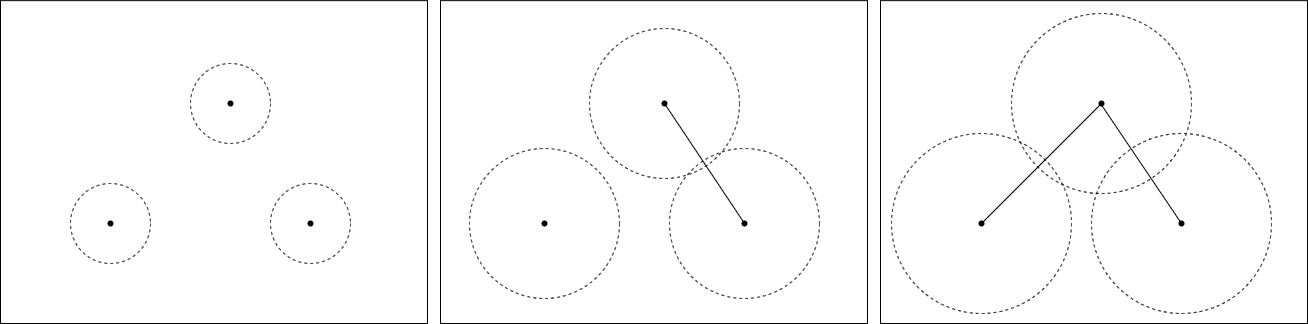}
    \caption{From left to right, Vietoris-Rips complex associated with the three points $x_1=(0,0)$, $x_2=(3,0)$ and $x_3=(2,2)$ for $\epsilon=0.5$, $\epsilon=2.5$, and $\epsilon=2.9$ respectively.}
    \label{fig:example2}
\end{figure}

\end{example}

In our case, we work with Vietoris-Rips filtrations that are determined by the value of $\epsilon$. It is easy to see that $VR_\epsilon(M)$ is a subcomples of  $VR_{\epsilon'}(M)$ for all $\epsilon'$ such that $0\leq\epsilon\leq\epsilon'$. In addition, since we work with a finite set $M$, there are only finitely many pairwise distances $d(x,y)$ among the elements of $M$, so there are only finitely many $\epsilon$ values where new simplices are added to $VR_\epsilon(M)$. Thus, as the radius $\epsilon$ increases, the Vietoris-Rips filtration of our dataset is built. This filtration allows us to study the topological features of a given dataset for different dimensions. These features, such as connected components, holes or voids, will be created and destroyed as the radius $\epsilon$ increases. By identifying the radius $\epsilon$ in the sequence when the topological features appear and disappear, we obtain a collection of birth and death pairs for each feature of each dimension. These pairs define the persistence diagram of a space.

\begin{definition}[Persistence diagram]
Let $(M,d)$ be a finite metric space, $\{\epsilon_0,\epsilon_1,\dots\epsilon_n\}$ be real numbers that verify $0\leq\epsilon_0<\epsilon_1<\dots\epsilon_n<\infty$ and 
\[
VR_{\epsilon_0}\subset VR_{\epsilon_1}\subset \cdots\subset VR_{\epsilon_{n-1}}\subset VR_{\epsilon_n}
\]
be the Vietoris-Rips filtration of $M$. Then we define the persistence diagram of M as 
\[
X=\{a_1,\dots, a_m\}
\]
where $a_i=(\epsilon_r,\epsilon_s)$ is the pair birth and death of a feature.
\end{definition}

\begin{example}
Let us consider the points $x_1=(0,0)$, $x_2=(3,0)$ and $x_3=(2,2)$ in the Euclidean space, and $VR_\epsilon$ the Vietoris-Rips filtration of these points. Then, the associated persistence diagram can be seen in Figure~\ref{fig:example3}.

\begin{figure}[ht]
\centering
	\includegraphics[width=0.6\textwidth]{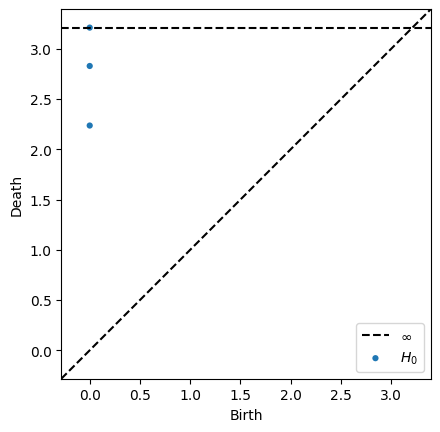}
    \caption{Persistence diagram of the points $x_1=(0,0)$, $x_2=(3,0)$ and $x_3=(2,2)$.}
    \label{fig:example3}
\end{figure}

\end{example}

Persistence diagram allows us to know the topological features of a space, and also allows us to compare two topological spaces. If two persistence diagrams are similar, we can conclude that their associated topological spaces are also similar. In order to conduct such a comparison, we need a metric that allows us to compare two persistence diagrams. Thus, the concept of distance between persistence diagrams arises. As we have seen, persistence diagrams are sets of points that determine the topological space; so, before talking about distance, we must define how we can establish a matching between two sets of points.

\begin{definition}[Matching]
Let $P$ and $Q$ be multisets in $\mathbb{R}^2$. We define a matching between $P$ and $Q$ to be a collection of pairs $X=\{(p,q)\in P\times Q\}$, where $p$ and $q$ can occur in at most one pair. If $(p,q)\in X$ then we say that $p$ is matched to $q$, otherwise if a given $p\in P$ does not belong to any pair in $X$, we say that $p$ is unmatched.
\end{definition}

Once we have defined a matching between two sets in $\mathbb{R}^2$, we can define a cost function that allows us to know how good is such a matching.

\begin{definition}[Cost]
Let $P$ and $Q$ be multisets in $\mathbb{R}$ with a matching $X$. Then we define a function $c: X \rightarrow \mathbb{R}^2$, called the cost, which maps a pair $(p,q)=((p_1,p_2),(q_1,q_2))$ to 
\[
((p_1,p_2),(q_1,q_2))\rightarrow max(\mid q_1-p_1\mid, \mid q_2-p_2\mid)
\]

We define the cost of a point $p=(p_1,p_2)$ to be
\[
c(p)= \frac{\mid p_2-p_1\mid}{2}
\]

Furthermore, we define the cost of the matching $X$ as
\[
c(X)=max(sup_{(p,q)\in X} c(p,q), sup_{ p\in P\cup Q, unmatched} c(p))
\]
\end{definition}

Finally, from these two concepts, we can introduce the notion of distance between persistence diagramas. In particular, we are going to use two different distances, the Bottleneck distance~\cite{bottleneck} and the Wasserstein distance~\cite{wasserstein}.

\begin{definition}[Bottleneck distance]
Let $P$ and $Q$ be multisets in $\mathbb{R}^2$. The Bottleneck distance between $P$ and $Q$ is defined as
\[
d_B(P,Q)= inf\{c(X)\mid X \text{ is a matching between } P\text{ and }Q\}
\]
\end{definition}

\begin{definition}[Wasserstein distance]
Let $P$ and $Q$ be multisets in $\mathbb{R}^2$. The r-Wasserstein distance between $P$ and $Q$ is defined as
\[
W_r(P,Q)= inf (\sum_{(x,y)\in X} \left\|q-p\right\|^{r}_\infty +\sum_{x\in X^c}\mid p_2-p_1\mid^r)^{\frac{1}{r}}
\]
where $X$ is a matching and $X^c$ is the set of unmatching points.
\end{definition}

After introducing these concepts, we provide a brief overview to the other area where this work can be framed: semi-supervised learning.

\subsection{Semi-supervised learning}
Semi-supervised methods take advantage of both labelled and unlabelled data~\cite{Laine17,Berthelot19}. These methods can be grouped into three main types: self-training, consistency regularisation and hybrid methods. In Self-Training methods a model is trained on labelled data and used to predict pseudo-labels for the unlabelled data. The model is then trained on both ground truth labels and pseudo-labels simultaneously. Some examples of these methods are pseudo label~\cite{pseudolabel} and noisy student~\cite{noisystudent}. Consistency regularisation methods, such as virtual adversarial training~\cite{virtualadversarial}, mean teacher~\cite{meanteachers} or $\pi$-models~\cite{pimodels}; use the idea that model predictions on an unlabelled image should remain the same even after adding some noise. Finally, hybrid methods combines ideas from self-training and consistency regularisation along with additional components for performance improvement. These methods include FixMatch~\cite{FixMatch} and MixMatch~\cite{MixMatch}.

Two of the most widely employed semi-supervised learning methods are Label Spreading~\cite{labelSpread} and Label Propagation~\cite{labelProp}. Both methods are based on label inference on unlabelled data using a graph-based approach. Label propagation computes a similarity matrix between samples and uses a KNN-based approach to propagate samples; whereas label spreading takes a similar approach but adds a regularisation step to be more robust to noise. 

% In the next section we will explain the methods used in this work to create a semi-supervised method using TDA.

\section{Methods}\label{sec:methods}
In this section, we describe the semi-supervised learning algorithms that we have designed to tackle binary classification tasks. In particular, we have studied two different approaches: a homological approach and a connectivity approach. In both of them, we start with a set $X_1$ of points from class 1, a set $X_2$ of points from class 2; and a set $X$ of unlabelled points. The objective of our algorithms is to annotate the elements of $X$ by using topological properties of $X_1$, and $X_2$.

\subsection{Homological method}
The first approach consists in studying the topological properties of the sets $X_1$ and $X_2$, and how those properties change when a new point $x\in X$ is added to each one of those sets. 
%Ejemplo

The hypothesis is that if a point belongs to a set, the topological variation that such a set will suffer when adding the point will be minimal; whereas if we add a point that does not belong to the set, the variation will be greater. In particular, we are going to calculate the persistence diagrams of each set and see how those diagrams vary when adding a new point. 

In particular, our semi-supervised learning algorithm takes as input the sets $X_1$ and $X_2$, a point $x\in X$, a threshold value $t$, and a flag that indicates whether the bottleneck or the Wasserstein distance should be used, we denote the chosen distance as $d$. The output produced by our algorithm is whether the point $x$ belongs to $X_1$, $X_2$ or none of them. In order to decide the output of the algorithm, our hypothesis is that if a point belongs to $X_1$, analogously for $X_2$, the topological variation that $X_1$ will suffer when adding the point will be minimal; whereas if the point does not belong to $X_1$, the variation will be greater. In particular, we proceed as follows:

\begin{enumerate}
    \item Construct the Vietoris-Rips filtrations $V_{X_1}$, $V_{X_2}$, $V_{X_1 \cup \{x\}}$ and $V_{X_2 \cup \{x\}}$;
    \item Construct the persistence diagrams $P(V_{X_1})$, $P(V_{X_2})$, $P(V_{X_1 \cup \{x\}})$ and $P(V_{X_2 \cup \{x\}})$;
    \item Compute the distances $d(P(V_{X_1}),P(V_{X_1 \cup \{x\}}))$ and $d(P(V_{X_2}),P(V_{X_2 \cup \{x\}}))$, from now on $d_1$ and $d_2$ respectively;
    \item If both $d_1$ and $d_2$ are greater than the threshold $t$, return none; otherwise, return the set associated with the minimum of the distances $d_1$ and $d_2$.
\end{enumerate}

The above algorithm is diagrammatically described in example~\ref{exam:example1}, and it is applied for all the points of the set of unlabelled points $X$. Note that if we use a threshold value of $0$, the algorithm will annotate all the points of $X$; however, this might introduce some noise as we will see in Section~\ref{sec:results}.

\begin{example}\label{exam:example1}
We take 9 points of the class 0, 10 points of the class 2, and 1 unlabelled points, as presented in Figure~\ref{fig:exampleHomolo}, and we apply our homological method.

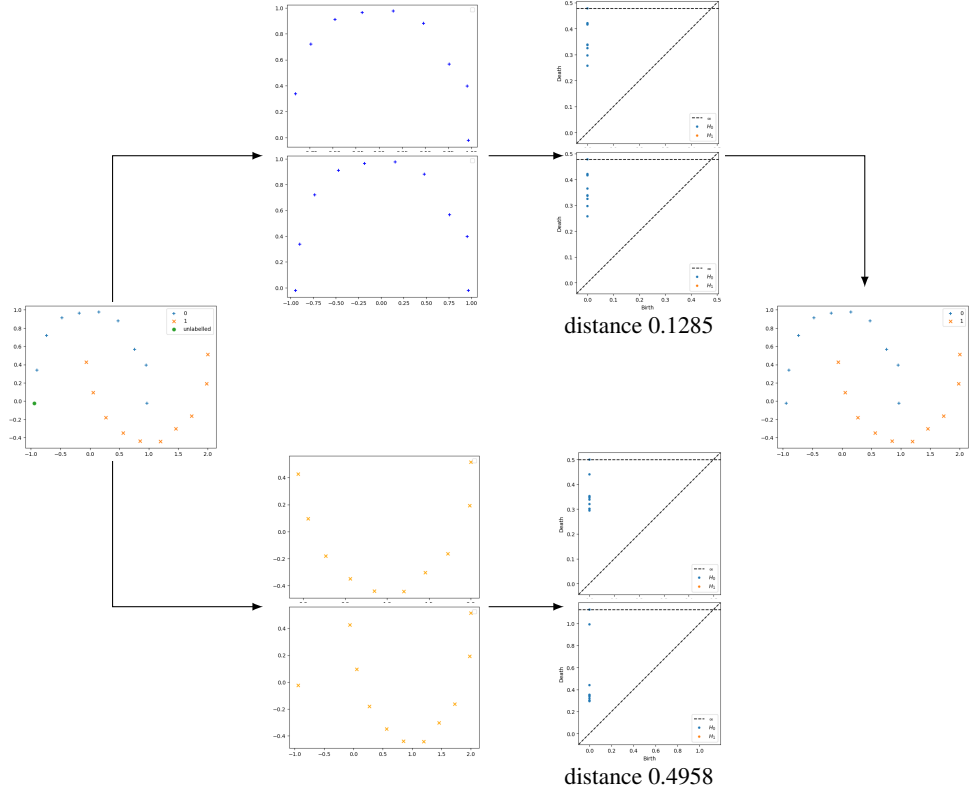
\begin{figure}[ht]
\centering
\begin{tikzpicture}
\draw[-latex] (0,0) -- (0,3) -- (2,3);
\draw[-latex] (0,0) -- (0,-3) -- (2,-3);
\draw[-latex] (8,3) -- (10,3) -- (10,1.25);
%\draw[-latex] (4,-3) -- (10,-3) -- (10,-1.25);
\draw (0,0) node{\includegraphics[scale=0.2]{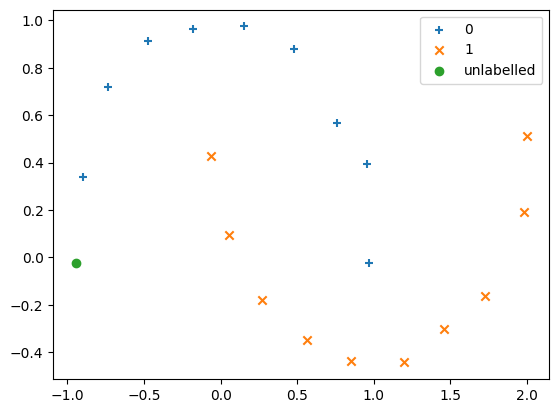}};

\draw (3.5,4) node{\includegraphics[scale=0.2]{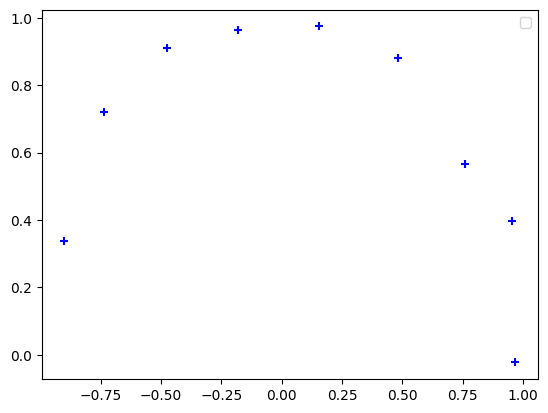}};
\draw (3.5,2) node{\includegraphics[scale=0.2]{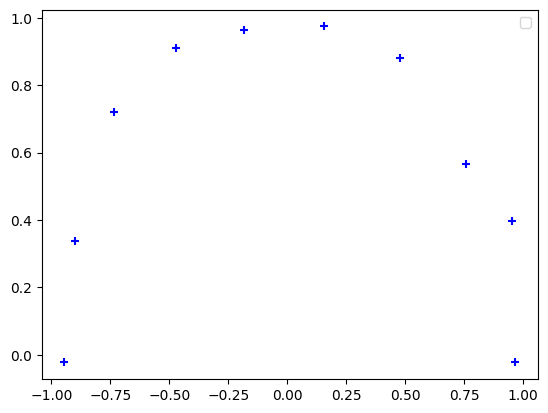}};
\draw (7,4) node{\includegraphics[scale=0.2]{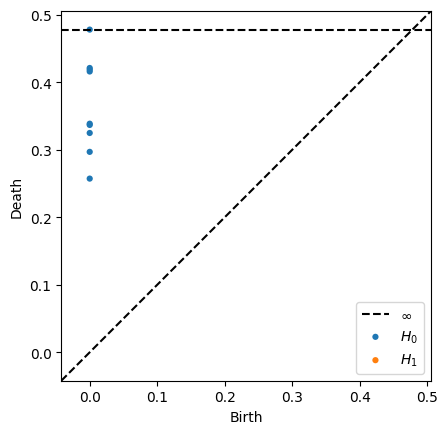}};
\draw (7,2) node{\includegraphics[scale=0.2]{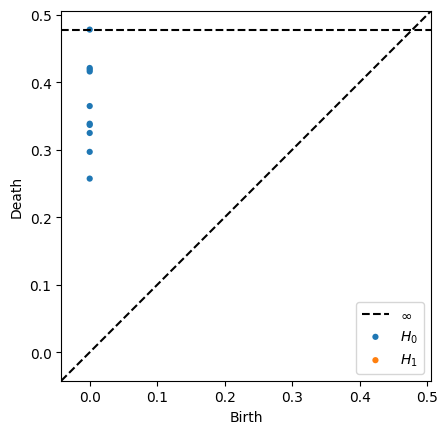}};
\draw[-latex] (5,3) -- (6,3);
\draw (7,0.75) node{{\small distance 0.1285}};

\draw (3.5,-2) node{\includegraphics[scale=0.2]{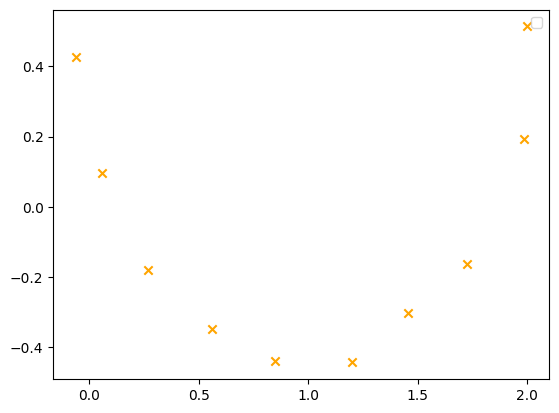}};
\draw (3.5,-4) node{\includegraphics[scale=0.2]{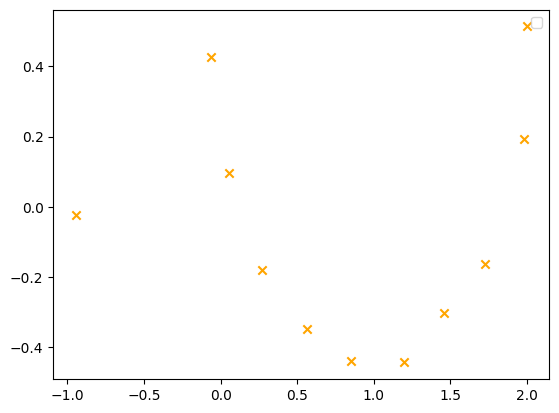}};
\draw (7,-2) node{\includegraphics[scale=0.2]{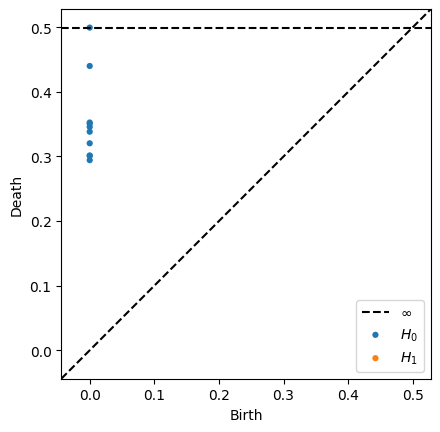}};
\draw (7,-4) node{\includegraphics[scale=0.2]{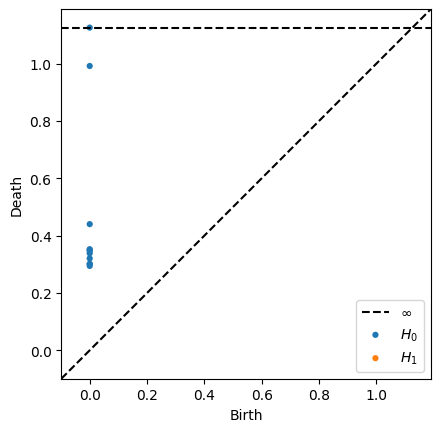}};
\draw[-latex] (5,-3) -- (6,-3);
\draw (7,-5.25) node{{\small distance 0.4958}};

\draw (10,0) node{\includegraphics[scale=0.2]{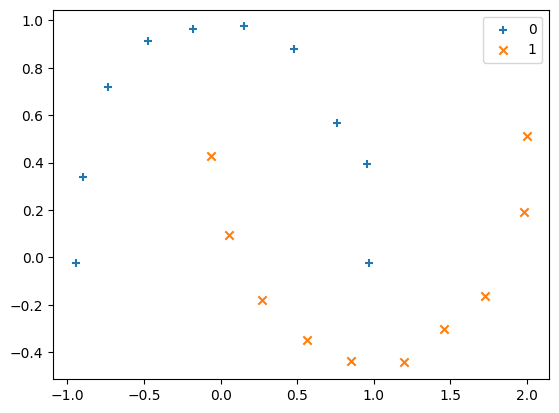}};

\end{tikzpicture}
    \caption{Example of the application of our homological method using the bottleneck distance, and using $0.6$ as threshold value.}
    \label{fig:exampleHomolo}
\end{figure}

The complete process can be seen on the project webpage.

% \begin{figure}[ht]
% \centering
% 	\includegraphics[width=0.6\textwidth]{ejemploAlgoritmo.png}
%     \caption{Example of 8 points of the class 0, 10 points of the class 1, and 2 unlabelled points}
%     \label{fig:exampleHomolo}
% \end{figure}

% If we apply our algorithm with a threshold of 0.6, the point $(-0.95, 0.0)$ would be labelled with class 0 while the point $(0.97, 0.0)$ would remain unlabelled. The complete process can be seen on the project webpage.
\end{example}

\subsection{Connectivity method}\label{sec:connectivity}
In the second method, we look at the connectivity of the data. In particular, we focus on the minimum radius that the Vietoris-Rips complex associated with a set has to take to be connected.As in the previous case, we start with a set $X_1$ of points from class 1, a set $X_2$ of points from class 2, and a set $X$ of unlabelled points. The objective of our algorithms is to annotate the elements of $X$ by using topological properties of $X_1$ and $X_2$. In particular, our semi-supervised learning algorithm takes as input the sets $X_1$ and $X_2$, a point $x\in X$. The output produced by our algorithm is whether the point $x$ belongs to $X_1$, $X_2$ or none of them. In order to decide the output of the algorithm, our hypothesis is that if a point belongs to $X_1$, analogously for $X_2$, the minimum connectivity radius of the associated Vietoris-Rips complex does not change considerably; on the contrary, if the point does not belong to the set $X_1$, analogously for $X_2$, the radius will increase. In particular, we proceed as follows:

\begin{enumerate}
    \item Construct the Vietoris-Rips complex $V_{X_1}$, $V_{X_2}$, $V_{X_1 \cup \{x\}}$ and $V_{X_2 \cup \{x\}}$;
    \item Compute the minimum connectivity radius $r(V_{X_1})$, $r(V_{X_2})$, $r(V_{X_1 \cup \{x\}})$ and $r(V_{X_2 \cup \{x\}})$, from now on $r_1$, $r_2$, $r_1^{\prime}$ and $r_2^{\prime}$ respectively;
    \item Compute the radius variation $\lvert r_1-r_1^{\prime} \rvert$ and $\lvert r_2-r_2^{\prime} \rvert$ from now on $d_1$ and $d_2$ respectively;
    \item If both $d_1$ and $d_2$ are zero, return none; otherwise, return the set associated with the minimum of the differences $d_1$ and $d_2$.
\end{enumerate}

In particular, to label the point with this method we have two variants. In the first case, we will say that a point belongs to a class if its radius has not been modified when adding it to that set; that is, if $d_i=0$, otherwise we will add it to the other set ($d_j\neq 0$). In the second case, we look at which radio has undergone the least variation ($d_i<d_j$) and we add it to that class; if the two variations are equal, we leave it unlabelled.

The above algorithm is diagrammatically described in example~\ref{exam:example2}, and it is applied for all the points of the set of unlabelled points $X$.
\begin{example}\label{exam:example2}
We take 9 points of the class 1, 10 points of the class 1, and 1 unlabelled points, as presented in Figure~\ref{fig:exampleConect}.
% \begin{figure}[ht]
% \centering
% 	\includegraphics[width=0.6\textwidth]{ejemploAlgoritmo.png}
%     \caption{Example of 8 points of the class 0, 10 points of the class 1, and 2 unlabelled points}
%     \label{fig:exampleConnect}
% \end{figure}

\begin{figure}[ht]
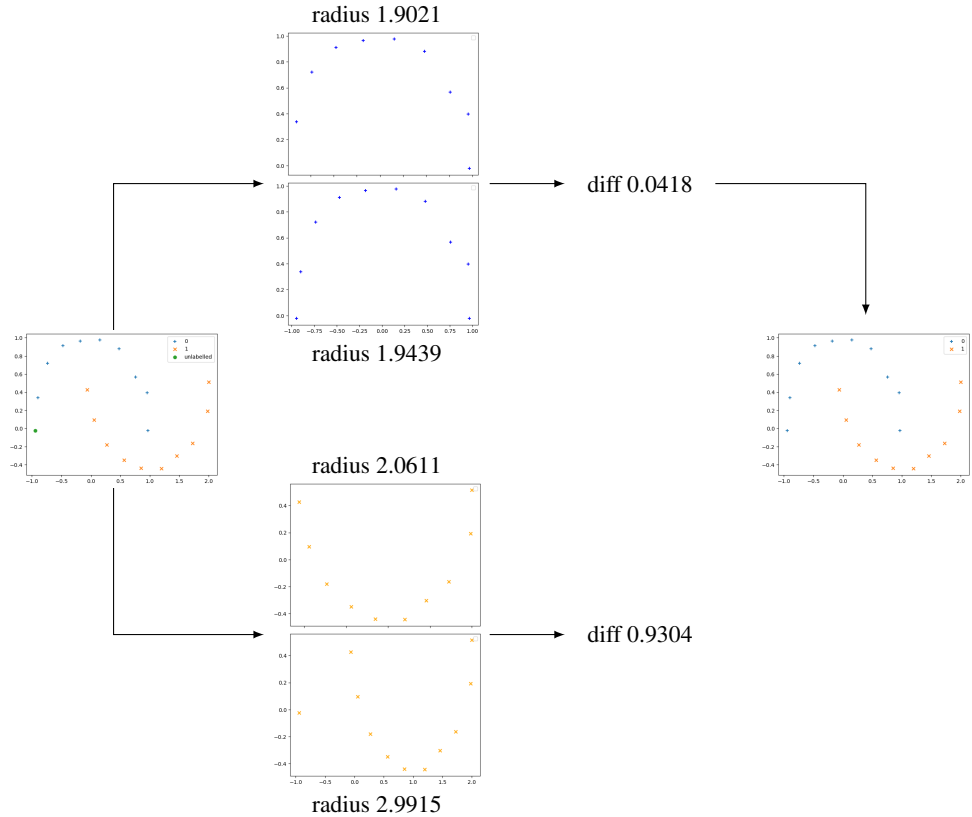

\centering
\begin{tikzpicture}
\draw[-latex] (0,0) -- (0,3) -- (2,3);
\draw[-latex] (0,0) -- (0,-3) -- (2,-3);
\draw[-latex] (8,3) -- (10,3) -- (10,1.25);
%\draw[-latex] (4,-3) -- (10,-3) -- (10,-1.25);
\draw (0,0) node{\includegraphics[scale=0.2]{Images/imagen1.png}};

\draw (3.5,4) node{\includegraphics[scale=0.2]{Images/imagen21.png}};
\draw (3.5,5.25) node{{\small radius 1.9021}};
\draw (3.5,2) node{\includegraphics[scale=0.2]{Images/imagen22.png}};
\draw (3.5,0.75) node{{\small radius 1.9439}};

\draw[-latex] (5,3) -- (6,3);
\draw (7,3) node{{\small diff 0.0418}};

\draw (3.5,-2) node{\includegraphics[scale=0.2]{Images/imagen31.png}};
\draw (3.5,-0.75) node{{\small radius 2.0611}};
\draw (3.5,-4) node{\includegraphics[scale=0.2]{Images/imagen32.png}};
\draw (3.5,-5.25) node{{\small radius 2.9915}};

\draw[-latex] (5,-3) -- (6,-3);
\draw (7,-3) node{{\small diff 0.9304}};

\draw (10,0) node{\includegraphics[scale=0.2]{Images/imagen4.png}};

\end{tikzpicture}
    \caption{Example of the application of our connectivity method.}
    \label{fig:exampleConect}
\end{figure}

% If we apply our algorithm with a threshold of 0.6, the point $(-0.95, 0.0)$ would remain unlabelled, while the point $(0.97, 0.0)$ would be labelled as 1. 

The complete process can be seen on the project webpage.

\end{example}

% \begin{figure}[t]
% \begin{lstlisting}[frame=single]
% Input:
%     $X_1$ set of points labelled with class 1.
%     $X_2$ set of points labelled with class 2.
%     $X_3$ set of unlabelled points.
%     $th$ threshold.

% Algorithm:
% $K_1$=Vietoris_Rips_Complex$(X_1)$
% $K_2$=Vietoris_Rips_Complex$(X_2)$

% $r_1$=radius$(K_1)$
% $r_2$=radius$(K_2)$

% for $x \in X_3$:
%     $X_1^{\prime}= X_1 \cup \{x\}$
%     $X_2^{\prime}= X_2 \cup \{x\}$
    
%     $K_1^{\prime}$=Vietoris_Rips_Complex$(X_1^{\prime})$
%     $K_2^{\prime}$=Vietoris_Rips_Complex$(X_2^{\prime})$

%     $r_1^{\prime}$=radius$(K_1^{\prime})$
%     $r_2^{\prime}$=radius$(K_2^{\prime})$
    
%     $d_1=abs(r_1^{\prime}-r_1)$
%     $d_2=abs(r_2^{\prime}-r_2)$
    
%     if $d_1<d_2$ then:
%         $Y_1=Y_1\cup \{x\}$
%     else if $d_1>d_2$ then:
%         $Y_2=Y_2\cup \{x\}$
%     else
%         $Y_3=Y_3\cup \{x\}$
%     endif
% endfor
% return $Y_1, Y_2, Y_3$
% \end{lstlisting}
% \caption{\textcolor{black}{Pseudocode of the algorithm 2.}}\label{fig:pseudocodeConnectivity2}
% \end{figure}

\subsection{API}
In order to facilitate the reproducibility of our methods, and also to simplify the application of the aforementioned semi-supervised learning algorithms to binary classification problems to other researchers, we have designed a Python library, available at the project webpage, that implements them. The library provides an API that is summarised in Figure~\ref{fig:API}. Several settings can be configured for the methods of the API, and we explain those options in the documentation of the project webpage. In order to employ, for instance, the homological method, the user must provide the annotated data, that is, a numpy array with all the annotated data, and a numpy array with the label of each data point; the unlabelled data in a numpy array format, the name of the distance to be used (Bottleneck or Wasserstein), the confidence threshold and if dimensionality reduction has to be applied. From that information, the library will automatically annotate the unlabelled data.

We have used the scikit-tda library~\cite{scikittda} to implement the homological distance method, whereas the Gudhi library~\cite{gudhi} has been used for implementing the connectivity method. All the methods have been implemented and tested with the Python programming language and using the Google Collaboratory environment~\cite{colab}.

\begin{figure}[ht]
\begin{lstlisting}[frame=single]
homologicalAnnotation(data, target, unlabelled_data, 
    distance, confidence, reduction)

connectivityAnnotation(data, target, unlabelled_data, 
    type, reduction)

\end{lstlisting}
\caption{API of the annotation methods provided in our library. The \texttt{data} is the labelled data in numpy array format. \texttt{target} are the labels of the data in numpy array format. The \texttt{unlabelled\_data} is the unlabelled data in numpy array format. The \texttt{distance} parameter refers to the name of the distance to use (bottleneck or Wasserstein). The \texttt{type} parameter is the type of condition to label a point in connectivity methods (0 or 1). The \texttt{Confidence} parameter refers to the confidence threshold. The \texttt{reduction} parameter denotes if dimensionality reduction is to be applied using the UMAP algorithm.}\label{fig:API}
\end{figure}

\section{Evaluation protocol}\label{sec:eval}
In this section, we present the datasets, the procedures and tools used for training and evaluating the methods explained in the previous section. We start by introducing the datasets that have been used for our experiments.

\subsection{Datasets}
In this work, we have used 10 different datasets that are summarised in Table~\ref{tab:datasets}. We have chosen datasets with different types of data; in particular, we have selected 3 synthetic datasets, 5 datasets of structured data, and 2 datasets of images. All datasets come from binary classification problems. We briefly describe each of the datasets below.

\begin{table}[h]
\centering
\resizebox{\linewidth}{!}{%
{\small

\begin{tabular}{ccccc}
 \toprule
  Dataset & \# Examples & \# Unlabelled examples & \# Features & Type\\
 \midrule
 \rowcolors{0}{white}{black!10!white}
 Blobs & 300 & 250 & 2 & Synthetic\\
 Circles & 300 & 250 & 2 & Synthetic\\
 Moons & 300 & 250 & 2 & Synthetic\\
 \midrule
 Banknote & 1372 & 1322 & 4 & Structured\\
 Breast Cancer & 569 & 519 & 30 & Structured\\
 Ionosphere & 351 & 301 & 34 & Structured\\
 Pima Indian Diabetes & 768 & 718 & 8 & Structured\\
 Sonar & 208 & 158 & 60 & Structured\\
 \midrule
 LiverGenderAL & 265 & 215 & 146688 & Images\\
 LiverGenderCR & 303 & 253 & 146688 & Images\\
 \bottomrule
\end{tabular}}}
\caption{Description of the datasets employed in our experiments.}\label{tab:datasets}
\end{table}

\begin{figure}[ht]
\begin{subfigure}[b]{0.49\textwidth}
    \includegraphics[width=\textwidth]{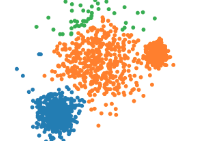}
    \label{fig:blobs}
  \end{subfigure}
  \hfill
    \begin{subfigure}[b]{0.49\textwidth}
    \includegraphics[width=\textwidth]{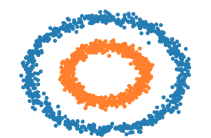}
    \label{fig:circles}
  \end{subfigure}
  \centering
  \begin{subfigure}[b]{0.49\textwidth}
    \includegraphics[width=\textwidth]{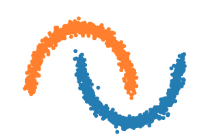}
    \label{fig:moons}
  \end{subfigure}
 \caption{Examples of the different datasets used. Top left Blobs dataset. Top right Circles dataset. Bottom Moons dataset.} \label{fig:datasets}
\end{figure}

The three synthetic datasets are generated using the scikit-learn library~\cite{scikit-learn} and contain 2D points. These datasets are the Blobs dataset, that consists of a normally-distributed cluster of points; the Circles dataset, that consists of two concentric circles one inside the other; and, the Moons dataset, whose points are distributed forming two interleaving half circles, see Figure~\ref{fig:datasets}. The structured datasets were taken from the UCI Machine Learning Repository~\cite{datasets}. The Banknote~\cite{datasets} dataset contains a number of measures taken from a photograph to predict whether a given banknote is authentic. The Breast Cancer~\cite{datasets} dataset is a structured dataset with 30 features that describe characteristics of the cell nuclei present in a digitized image of a Fine Needle Aspirate (FNA) of a breast mass. The Ionosphere~\cite{datasets} dataset is designed to predict the structure in the atmosphere given radar returns targeting free electrons in the ionosphere. The Pima Indians Diabetes~\cite{pimaIndian} dataset involves predicting the onset of diabetes within 5 years in Pima Indians given medical details. The Sonar~\cite{datasets} dataset involves the prediction of whether or not an object is a mine or a rock given the strength of sonar returns at different angles. Finally, for the image datasets, we have used the two Liver Gender datasets~\cite{livergender} which feature microscopy images of tissue, from both men and women. 

%All the elements of the datasets will be converted into data vectors, for synthetic datasets, each element in the dataset will be a vector whose elements will be the coordinates $x$ and $y$ of the points. For structured data, each feature will be a component of the generated vector. For image datasets, the elements of the generated vectors will be the pixels of each image.

For our study, we have split each of the datasets of the benchmark into two different sets: a training set with the $80 \%$ of the data, and a testing set with the $20 \%$ of the data, except for LiverGenderAL and LiverGenderCR datasets that we use the existing split provided by the dataset. In addition, for each training dataset, we have selected $25$ samples per class using them as labelled data, and removing the annotation of the rest of the training data to test the semi-supervised learning methods. 

% The splits used in our experiments and more information about datasets are available in the project webpage.

\subsection{Training and evaluation procedure}
To check the correct performance of our methods we have trained two classic machine learning algorithms that are SVM~\cite{svm} and Random Forest~\cite{randomForest} using the scikit-learn functionality~\cite{scikit-learn}. In particular, we have trained these models with the initial annotated data obtaining a base result. Subsequently we have used the developed methods, and three classical semi-supervised learning methods (Label Propagation, Label Spreading~\cite{labelSpread}, and self-training~\cite{selfTrainClas}) to annotate the unlabelled data. Finally, we have retrained the two ML models with all the annotated data, to see the variation in performance of the models. Such a performance of the models has been evaluated using the accuracy. In addition, in order to evaluate the behaviour of the annotation methods we have taken into account the percentage of the data points correctly labelled and the percentage of data labelled with respect to the total available data. 

For testing our methods, we have used 20 variations of the homogical method and 4 variations of the connectivity method. For the homological variants, half of them use the Bottleneck distance, and the other half use the Wasserstein distance. In addition, for both distances we have established 5 different threshold levels (0.8, 0.6, 0.4, 0.2 and 0.0), except for the syntethic datasets that we have established only two threshold levels (0 and 0.8). Furthermore, for each of these variants, we have made two different versions, the former works with the original data points; whereas, in the latter we have reduced the dimensionality of the data to two dimension by using the UMAP reduction algorithm~\cite{umap}. In the case of the connectivity alternatives, we have used the two variants explained in Section~\ref{sec:connectivity}, we will called these methods connectivity1 and connectivity2. In addition, as in the case of homological methods, we have considered a version with the original data and another where we have reduced the dimensionality of the data to two dimensions using UMAP.

\section{Results and discussion}\label{sec:results}

In this section, we present a thorough analysis for the results obtained by the developed semi-supervised methods and the 3 traditional semi-supervised methods. Due to the nature of the data in each of the datasets, we have decided to separate our study into three different groups. First, we have studied the performance of the semi-supervised learning methods when applied to synthetic datasets; then, when applied to the structured datasets; and finally, to the image datasets.

Table~\ref{tab:synthetics} includes the behaviour of the semi-supervised learning methods when applied to the three synthetic datasets (Blobs, Circles, and Moons). The results show that there are no major differences between the semi-supervised learning methods, although we can observe that the homological method with the Wasserstein distance as well as the connectivity methods offer slightly worse results than the rest.

The performance of our methods on structured datasets is included in Table~\ref{tab:structured}. From these results we can withdraw some conclusions. In general, connectivity methods do not offer good results; in fact, despite having a good annotation accuracy in many cases, the number of annotations they obtain is quite low. In addition, they obtain worse results that the base classifiers (see Appendix~\ref{app:tables}). On the contrary, homological methods offer good results, improving the base results in most cases, see Table~\ref{tab:structured}. These methods work better when a confidence threshold is set. Namely, we have established threshold values of 0.8, 0.6, 0.4, and 0.2; and we have observed that from a threshold value higher than 0.4, all the unlabelled data points are labelled. Furthermore, the best results are obtained when we obtain a balance between the amount of annotated data and the correctness of this labelled data. In particular, the best results are obtained with a threshold of 0.8. Regarding the distance to be used, there are not significant differences between the Bottleneck and the Wasserstein distance. 

We have also studied what happens when we reduce the dimensionality of the data by using UMAP. In view of the results, we can see that although there are no great differences, in general the results of the SVM and RF classifiers decrease slightly. This may be because despite the amount of annotated data increases, the correctness of this labelled data decreases. Therefore, we can conclude that the best results are obtained with the homological method using the Wasserstein distance, although there are no major differences with using the bottleneck distance, applying a threshold of 0.8 and without using the reduction of dimension. This method, in general, improves the results obtained in the base case and even improves the results obtained by the 3 classic annotation methods.

\begin{table*}[h]
\centering
\resizebox{\linewidth}{!}{%
{\small

\begin{tabular}{c|cc|cc|cc|cc|cc|cc}
 \toprule
   & \multicolumn{2}{c|}{Banknote} & \multicolumn{2}{c|}{Breast Cancer}  & \multicolumn{2}{c|}{Ionosphere} & \multicolumn{2}{c|}{Prima Indian} & \multicolumn{2}{c}{Sonar} & \multicolumn{2}{c}{Mean(STD)}\\
  Method & SVM & RF & SVM & RF & SVM & RF & SVM & RF & SVM & RF & SVM & RF\\
 \midrule
 \rowcolors{0}{white}{black!10!white}
 Base & 97.0 & 88.6 & 89.3 & \textbf{96.1} & 80.0 & 93.3 & 65.7 & 60.8 & 61.3 & 64.5 & 78.7(15.2) & 80.7(16.7)\\
 \midrule
 Label Propagation & 97.4 & 93.2 & 90.3 & 89.3 & 86.7 & 86.7 & 64.3 & 68.5 & 58.1 & 54.8 & 79.3(17.1) & 78.5(16.3)\\
 Label Spreading & 97.4 & 93.2 & 90.3 & 89.3 & 86.7 & 86.7 & 64.3 & 68.5 & 58.1 & 54.8 & 79.3(17.1) & 78.5(16.3)\\
 Self Training classifier & 95.1 & 93.6 & 35.9 & 35.9 & 85.0 & 86.7 & 66.4 & 66.4 & 58.1 & 67.7 & 68.1(23.2) & 70.1(22.4)\\
 \midrule
 Bottleneck & 97.4 & 90.5 & 87.4 & 85.4 & 78.3 & 86.7 & 63.6 & 62.9 & 45.2 & 45.2 & 77.1(22.6) & 74.1(19.5)\\
 Bottleneck threshold 0.8 & \textbf{99.2} & 92.4 & 93.2 & 91.3 & 78.3 & 95.0 & 63.6 & 64.3 & 61.3 & 64.5 & 79.1(17.0) & \textbf{81.5(15.6)}\\
 Bottleneck threshold 0.6 & \textbf{99.2} & 91.3 & 89.3 & 90.3 & 75.0 & 88.3 & 59.4 & 63.6 & 48.4 & 45.2 & 74.3(20.9) & 75.7(20.6)\\
 Bottleneck threshold 0.4 & 97.4 & 90.5 & 87.4 & 85.4 & 78.3 & 86.7 & 63.6 & 62.9 & 45.2 & 45.2 & 74.4(20.5) & 74.1(19.5)\\
 Bottleneck threshold 0.2 & 97.4 & 90.5 & 87.4 & 85.4 & 78.3 & 86.7 & 63.6 & 62.9 & 45.2 & 45.2 & 74.4(20.5) & 74.1(19.5)\\
 Bottleneck UMAP & 97.4 & 96.2 & 92.2 & 92.2 & 85.0 & 88.3 & 58.7 & 53.2 & 67.7 & 64.5 & \textbf{80.2(16.4)} & 78.9(18.9)\\
 Bottleneck UMAP threshold 0.8 & 97.4 & 94.3 & 91.3 & 88.4 & 86.7 & 93.3 & 56.6 & 59.4 & 64.5 & 61.3 & 79.3(17.7) & 79.3(17.5)\\
 Bottleneck UMAP threshold 0.6 & 97.4 & 94.6 & 91.3 & 90.3 & 86.7 & 90.0 & 57.3 & 57.3 & 67.7 & 71.0 & 80.1(16.9) & 80.6(15.9)\\
 Bottleneck UMAP threshold 0.4 & 97.4 & 96.2 & 92.2 & 92.2 & 85.0 & 88.3 & 58.7 & 53.2 & 67.7 & 64.5 & \textbf{80.2(16.4)} & 78.9(18.9)\\
 Bottleneck UMAP threshold 0.2 & 97.4 & 96.2 & 92.2 & 92.2 & 85.0 & 88.3 & 58.7 & 53.2 & 67.7 & 64.5 & \textbf{80.2(16.4)} & 78.9(18.9)\\
 Wasserstein & 97.0 & 96.2 & 87.4 & 87.4 & 76.7 & 81.7 & 60.8 & 62.9 & 71.0 & 71.0 & 78.6(14.1) & 79.8(13.2)\\
 Wasserstein threshold 0.8 & 97.4 & 89.8 & 92.2 & 88.4 & 80.0 & 95.0 & \textbf{68.5} & 67.8 & 61.3 & 64.5 & 79.9(15.3) & 81.1(13.9)\\
 Wasserstein threshold 0.6 & \textbf{99.2} & 93.6 & 89.3 & 87.4 & 70.0 & 91.7 & 61.5 & 61.5 & \textbf{74.2} & 61.3 & 78.9(15.2) & 79.1(16.3)\\
 Wasserstein threshold 0.4 & 97.0 & 96.2 & 87.4 & 87.4 & 76.7 & 81.7 & 60.8 & 62.9 & 71.0 & 71.0 & 78.6(14.1) & 79.8(13.2)\\
 Wasserstein threshold 0.2 & 97.0 & 96.2 & 87.4 & 87.4 & 76.7 & 81.7 & 60.8 & 62.9 & 71.0 & 71.0 & 78.6(14.1) & 79.8(13.2)\\
 Wasserstein UMAP & 97.0 & 95.8 & 92.2 & 91.3 & 78.3 & 91.7 & 58.0 & 57.3 & 71.0 & 67.7 & 79.3(15.8) & 80.8(17.1)\\
 Wasserstein UMAP threshold 0.8 & 97.4 & 95.1 & 91.3 & 87.4 & 85.0 & 93.3 & 57.3 & 63.6 & 64.5 & 67.7 & 79.1(17.3) & 81.4(14.7)\\
 Wasserstein UMAP threshold 0.6 & 97.0 & 95.5 & 94.2 & 91.3 & 86.7 & 90.0 & 58.7 & 56.6 & 64.5 & 67.7 & \textbf{80.2(17.5)} & 80.2(17.0)\\
 Wasserstein UMAP threshold 0.4 & 97.0 & 95.8 & 92.2 & 91.3 & 78.3 & 91.7 & 58.0 & 57.3 & 71.0 & 67.7 & 79.3(15.8) & 80.8(17.1)\\
 Wasserstein UMAP threshold 0.2 & 97.0 & 95.8 & 92.2 & 91.3 & 78.3 & 91.7 & 58.0 & 57.3 & 71.0 & 67.7 & 79.3(15.8) & 80.8(17.1)\\
 \midrule
 Connectivity1 & 93.6 & 87.9 & 89.3 & 93.2 & 76.7 & 88.3 & 61.5 & 62.9 & 64.5 & 61.3 & 77.1(14.3) & 78.7(15.3)\\
 Connectivity1 UMAP & 80.3 & 85.2 & 84.5 & 89.3 & 83.3 & \textbf{100} & 51.1 & 53.2 & 67.7 & 58.1 & 73.4(14.1) & 77.2(20.5)\\
 Connectivity2 & 93.6 & 87.5 & 89.3 & 93.2 & 71.7 & 83.3 & 60.1 & 58.7 & 64.5 & 64.5 & 75.8(14.9) & 77.5(15.0)\\
 Connectivity2 UMAP & 53.8 & 75.0 & 84.5 & 89.3 & 80.0 & 95.0 & 51.1 & 51.1 & 67.7 & 58.1 & 67.4(15.0) & 73.7(19.1)\\
 \bottomrule
\end{tabular}}}
\caption{Accuracy results for the SVM and RF classifiers trained with data annotated by each of the annotation methods (classical, homological and connectivity) together with the results obtained with the initial data (base) in the 5 structured datasets. Best results for each dataset are highlighted in bold face.}\label{tab:structured}
\end{table*}

Finally, we have studied the case of image datasets, see Table~\ref{tab:images}. In this case, the connectivity methods again perform poorly, worsening even the base results. It is remarkable the case of the LiverGenderAL dataset in which the annotation accuracy is quite high (around 95\%), while the annotation percentage is around 40\% and however the results of the classifiers are worse than in the base case. Homological methods offer good results in general, improving the baseline results. In this case, we can see a difference in performance when select a distance, since the results obtained using the Wasserstein distance exceed those obtained with the Bottleneck distance. When applying dimension reduction on these datasets, there is a big difference depending on the distance used. In the case of the Bottleneck distance, the performance improvement is notable when applying the dimensionality reduction. In particular, it greatly increases the annotation accuracy and the number of annotated images is maintained or increased. In the case of the Wasserstein distance, the performance with the reduction of dimensionality is very similar and even in some cases decreases. Another difference that we can observe is the results obtained when using different thresholds. In this case, the threshold of 0.8 means that no data is labeled when we do not apply dimensionality reduction, while the threshold of 0.4 and 0.2 label all the data, that is, they have the same performance as not using a threshold. Setting the threshold of 0.6 does not produce improvements. Therefore, we can conclude that in general the method that works best is the homological method using the Wasserstein distance without threshold. This method improves the base results in both datasets, both for the SVM classifier and the RF classifier. Also, in the LiverGenderAL dataset it outperforms classic annotation methods by more than 16\% and it obtains same results in the LiverGenderCR dataset.

\begin{table*}[ht]
\centering
\resizebox{\linewidth}{!}{%
{\small

\begin{tabular}{c|cc|cc|cc}
 \toprule
   & \multicolumn{2}{c|}{Liver Gender AL} & \multicolumn{2}{c}{Liver Gender CR} & \multicolumn{2}{c}{Mean(STD)}\\
   Method & SVM & RF & SVM & RF & SVM & RF\\
 \midrule
 \rowcolors{0}{white}{black!10!white}
 Base & 71.6 & 70.2 & 80.3 & 78.9 & 76.0(6.1) & 74.5(6.2)\\
 \midrule
 Label Propagation & 62.7 & 62.7 & 82.9 & 82.9 & 72.8(14.3) & 72.8(14.3) \\
 Label Spreading & 62.7 & 62.7 & 82.9 & 82.9 & 72.8(14.3) & 72.8(14.3)\\
 Self Training classifier & 52.2 & 52.2 & 52.9 & 57.9 & 55.7(4.9) & 55.1(4.0)\\
 \midrule
 Bottleneck & 47.8 & 47.8 & 53.9 & 53.9 & 50.9(4.4) & 50.9(4.4)\\
 Bottleneck threshold 0.8 & 70.2 & 64.2 & 80.3 & 78.9 & 75.2(7.1) & 71.6(10.4)\\
 Bottleneck threshold 0.6 & 70.2 & 64.2 & 80.3 & 78.9 & 75.2(7.1) & 71.6(10.4) \\
 Bottleneck threshold 0.4 & 47.8 & 47.8 & 54.0 & 54.0 & 50.9(4.4) & 50.9(4.4)\\
 Bottleneck threshold 0.2 & 47.8 & 47.8 & 54.0 & 54.0 & 50.9(4.4) & 50.9(4.4)\\
 Bottleneck UMAP & 71.6 & 68.7 & 80.3 & 80.3 & 76.0(6.1) & 74.5(8.2)\\
 Bottleneck UMAP threshold 0.8 & 70.2 & 64.2 & 81.6 & 80.3 & 75.9(8.1) & 72.2(11.4)\\
 Bottleneck UMAP threshold 0.6 & 74.6 & \textbf{79.1} & 81.6 & 80.3 & \textbf{78.1(4.9)} & 79.7(0.8)\\
 Bottleneck UMAP threshold 0.4 & 71.6 & 68.7 & 80.3 & 80.3 & 76.0(6.1) & 74.5(8.2)\\
 Bottleneck UMAP threshold 0.2 & 71.6 & 68.7 & 80.3 & 80.3 & 76.0(6.1) & 74.5(8.2)\\
 Wasserstein & 73.1 & 77.6 & 82.9 & 82.9 & 78.0(6.9) & \textbf{80.3(3.7)}\\
 Wasserstein threshold 0.8 & 70.2 & 64.2 & 80.3 & 78.9 & 75.2(7.1) & 71.6(10.4)\\
 Wasserstein threshold 0.6 & 71.6 & 68.7 & 82.9 & 82.9 & 77.3(8.0) & 75.8(10.1)\\
 Wasserstein threshold 0.4 & 73.1 & 77.6 & 82.9 & 82.9 & 78.0(6.9) & \textbf{80.3(3.7)}\\
 Wasserstein threshold 0.2 & 73.1 & 77.6 & 82.9 & 82.9 & 78.0(6.9) & \textbf{80.3(3.7)}\\
 Wasserstein UMAP & 74.6 & 73.1 & 65.8 & \textbf{86.8} & 70.2(6.2) & 80.0(9.7)\\
 Wasserstein UMAP threshold 0.8 & 68.7 & 59.7 & 82.9 & 76.3 & 75.8(10.1) & 68.0(11.8)\\
 Wasserstein UMAP threshold 0.6 & 73.1 & 65.7 & 81.6 & 79.0 & 77.4(6.0) & 72.3(9.4)\\
 Wasserstein UMAP threshold 0.4 & 74.6 & 73.1 & 65.8 & \textbf{86.8} & 70.2(6.3) & 80.0(9.7)\\
 Wasserstein UMAP threshold 0.2 & 74.6 & 73.1 & 65.8 & \textbf{86.8} & 70.2(6.3) & 80.0(9.7)\\
 \midrule
 Connectivity1 & 70.2 & 61.2 & 46.1 & 46.1 & 58.1(17.0) & 53.6(10.7)\\
 Connectivity1 UMAP & 71.6 & 59.7 & 75.0 & 73.7 & 73.3(2.4) & 66.7(9.9)\\
 Connectivity2 & 70.2 & 59.7 & 46.1 & 46.1 & 58.1(17.0) & 52.9(9.7)\\
 Connectivity2 UMAP & 70.2 & 64.2 & 76.3 & 72.4 & 73.2(4.4) & 68.3(5.8)\\
 \bottomrule
\end{tabular}}}
\caption{Accuracy results for the SVM and RF classifiers trained with data annotated by each of the annotation methods (classical, homological and connectivity) together with the results obtained with the initial data (base) in the 2 datasets of images. Best results for each dataset are highlighted in bold face.}\label{tab:images}
\end{table*}

\section{Conclusions and further work}
In this work, we have studied the combination of Topological Data Analysis techniques with semi-supervised learning methods to tackle binary classification problems with a limited amount of labelled data. The results show that the combination of these methods can create classification models that achieve better results than those obtained when using classical semi-supervised learning methods on different kinds of datasets. Specifically, the homological method developed using the Wasserstein distance with a threshold of 0.8 in the case of structured datasets, and without a threshold in image datasets, generally obtains the best results by improving, in some cases, more than a $16\%$ the results obtained by the classical semi-supervised learning methods.

In the future, we plan to study an iterative version of these methods and the application of ensemble techniques in order to improve the robustness and reliability of our methods. Finally, we plan to extend our work to non-binary classification problems.

\section{Acknowledgments}
This work was partially supported by Ministerio de Ciencia e Innovación [PID2020-115225RB-I00 / AEI / 10.13039/501100011033], and FPU Grant 16/06903 of the Spanish MEC.

\bibliographystyle{splncs04}
\bibliography{biblio}

\begin{thebibliography}{10}
\providecommand{\url}[1]{\texttt{#1}}
\providecommand{\urlprefix}{URL }
\providecommand{\doi}[1]{https://doi.org/#1}

\bibitem{Affonso17}
Affonso, C., Rossi, A.L.D., Vieira, F.H.A., et~al.: Deep learning for
  biological image classification. Expert Systems with Applications
  \textbf{85}(1),  114--122 (2017)

\bibitem{security}
{Akçay}, S., Kundegorski, M.E., Devereux, M., et~al.: Transfer learning using
  convolutional neural networks for object classification within x-ray baggage
  security imagery. In: 2016 IEEE International Conference on Image Processing.
  pp. 1057--1061. ICIP'16 (2016)

\bibitem{breastCancer}
Ara\'ujo, T., Aresta, G., Castro, E., et~al.: Classification of breast cancer
  histology images using convolutional neural networks. PLoS ONE
  \textbf{12}(6) (2017)

\bibitem{Berthelot19}
Berthelot, D., et~al.: Mixmatch: A holistic approach to semi-supervised
  learning. In: Advances in Neural Information Processing Systems 32, pp.
  5049--5059. Curran Associates, Inc. (2019)

\bibitem{MixMatch}
Berthelot, D., et~al.: Mixmatch: A holistic approach to semi-supervised
  learning. In: 33rd International Conference on Neural Information Processing
  Systems (NEURIPS'19). pp. 5050--5060. Curran Associates Inc. (2019)

\bibitem{colab}
Bisong, E.: Google Colaboratory, pp. 59--64 (2019).
  \doi{10.1007/978-1-4842-4470-8\_7},
  \url{https://doi.org/10.1007/978-1-4842-4470-8\_7}

\bibitem{svm}
Cortes, C., Vapnik, V.: Support-vector networks. Machine learning
  \textbf{20}(3),  273--297 (1995)

\bibitem{datasets}
Dua, D., Graff, C.: {UCI} machine learning repository (2017),
  \url{http://archive.ics.uci.edu/ml}

\bibitem{computationalTop}
Edelsbrunner, H., Harer, J.: Computational topology: an introduction. American
  Mathematical Soc. (2010)

\bibitem{persist}
Edelsbrunner, H., Letscher, D., Zomorodian, A.: Topological persistence and
  simplification. In: Proceedings 41st annual symposium on foundations of
  computer science. pp. 454--463. IEEE (2000)

\bibitem{bottleneck}
Efrat, A., Itai, A., Katz, M.J.: Geometry helps in bottleneck matching and
  related problems. Algorithmica  \textbf{31},  1--28 (2001)

\bibitem{fefferman2016testing}
Fefferman, C., Mitter, S., Narayanan, H.: Testing the manifold hypothesis.
  Journal of the American Mathematical Society  \textbf{29}(4),  983--1049
  (2016)

\bibitem{randomForest}
Ho, T.K.: Random decision forests. In: Proceedings of 3rd international
  conference on document analysis and recognition. vol.~1, pp. 278--282. IEEE
  (1995)

\bibitem{anotationDem}
Irvin, J., Rajpurkar, P., Ko, M., Yu, Y., Ciurea-Ilcus, S., Chute, C.,
  Marklund, H., Haghgoo, B., Ball, R., Shpanskaya, K., et~al.: Chexpert: A
  large chest radiograph dataset with uncertainty labels and expert comparison.
  In: Proceedings of the AAAI conference on artificial intelligence. vol.~33,
  pp. 590--597 (2019)

\bibitem{wasserstein}
Kantorovich, L.V.: Mathematical methods of organizing and planning production.
  Management Science  \textbf{6},  366--422 (1960). \doi{10.1287/mnsc.6.4.366},
  \url{https://doi.org/10.1287/mnsc.6.4.366}

\bibitem{Laine17}
Laine, S., Aila, T.: {Temporal Ensembling for Semi-Supervised Learning}. In:
  5th International Conference on Learning Representations. pp. 1--13. ICLR'17
  (2017)

\bibitem{pseudolabel}
Lee, D.H., et~al.: Pseudo-label: The simple and efficient semi-supervised
  learning method for deep neural networks. In: Workshop on challenges in
  representation learning, ICML. vol.~3, p.~896 (2013)

\bibitem{umap}
McInnes, L., Healy, J., Saul, N., Grossberger, L.: Umap: Uniform manifold
  approximation and projection. The Journal of Open Source Software
  \textbf{3}(29), ~861 (2018)

\bibitem{virtualadversarial}
Miyato, T., Maeda, S.i., Koyama, M., Ishii, S.: Virtual adversarial training: a
  regularization method for supervised and semi-supervised learning. IEEE
  transactions on pattern analysis and machine intelligence  \textbf{41}(8),
  1979--1993 (2018)

\bibitem{atmosphericTDA}
Muszynski, G., Kashinath, K., Kurlin, V., Wehner, M., et~al.: Topological data
  analysis and machine learning for recognizing atmospheric river patterns in
  large climate datasets. Geoscientific Model Development  \textbf{12}(2),
  613--628 (2019)

\bibitem{scikit-learn}
Pedregosa, F., Varoquaux, G., Gramfort, A., Michel, V., Thirion, B., Grisel,
  O., Blondel, M., Prettenhofer, P., Weiss, R., Dubourg, V., Vanderplas, J.,
  Passos, A., Cournapeau, D., Brucher, M., Perrot, M., Duchesnay, E.:
  Scikit-learn: Machine learning in {P}ython. Journal of Machine Learning
  Research  \textbf{12},  2825--2830 (2011)

\bibitem{brainTDA}
Saggar, M., Sporns, O., Gonzalez-Castillo, J., Bandettini, P.A., Carlsson, G.,
  Glover, G., Reiss, A.L.: Towards a new approach to reveal dynamical
  organization of the brain using topological data analysis. Nature
  communications  \textbf{9}(1),  1--14 (2018)

\bibitem{pimodels}
Samuli, L., Timo, A.: Temporal ensembling for semi-supervised learning. In:
  International Conference on Learning Representations (ICLR). vol.~4, p.~6
  (2017)

\bibitem{scikittda}
Saul, N., Tralie, C.: Scikit-tda: Topological data analysis for python (2019).
  \doi{10.5281/zenodo.2533369}, \url{https://doi.org/10.5281/zenodo.2533369}

\bibitem{livergender}
Shamir, L., Orlov, N., Eckley, D.M., et~al.: Iicbu 2008: A proposed benchmark
  suite for biological image analysis. Medical \& Biological Engineering \&
  Computing  \textbf{46}(9),  943--947 (2008)

\bibitem{pimaIndian}
Smith, J.W., Everhart, J.E., Dickson, W.C., Knowler, W.C., Johannes, R.S.:
  Using the adap learning algorithm to forecast the onset of diabetes mellitus.
  In: Proceedings of the Symposium on Computer Applications and Medical Care.
  pp. 261--265. IEEE (1988)

\bibitem{FixMatch}
Sohn, K., et~al.: Fixmatch: Simplifying semi-supervised learningwith
  consistency and confidence. In: 34th International Conference on Neural
  Information Processing Systems (NEURIPS'20). Curran Associates Inc. (2020)

\bibitem{dataDem}
Sun, C., Shrivastava, A., Singh, S., Gupta, A.: Revisiting unreasonable
  effectiveness of data in deep learning era. In: Proceedings of the IEEE
  international conference on computer vision. pp. 843--852 (2017)

\bibitem{meanteachers}
Tarvainen, A., Valpola, H.: Mean teachers are better role models:
  Weight-averaged consistency targets improve semi-supervised deep learning
  results. arXiv preprint arXiv:1703.01780  (2017)

\bibitem{networksTDA}
Taylor, D., Klimm, F., Harrington, H.A., Kram{\'a}r, M., Mischaikow, K.,
  Porter, M.A., Mucha, P.J.: Topological data analysis of contagion maps for
  examining spreading processes on networks. Nature communications
  \textbf{6}(1),  1--11 (2015)

\bibitem{gudhi}
{The GUDHI Project}: {GUDHI} User and Reference Manual. {GUDHI Editorial Board}
  (2015), \url{http://gudhi.gforge.inria.fr/doc/latest/}

\bibitem{noisystudent}
Xie, Q., Luong, M.T., Hovy, E., Le, Q.V.: Self-training with noisy student
  improves imagenet classification. In: Proceedings of the IEEE/CVF Conference
  on Computer Vision and Pattern Recognition. pp. 10687--10698 (2020)

\bibitem{selfTrainClas}
Yarowsky, D.: Unsupervised word sense disambiguation rivaling supervised
  methods. In: 33rd annual meeting of the association for computational
  linguistics. pp. 189--196 (1995)

\bibitem{labelSpread}
Zhou, D., Bousquet, O., Lal, T.N., Weston, J., Schölkopf, B.: Learning with
  local and global consistency. In: Advances in Neural Information Processing
  Systems 16. pp. 321--328. MIT Press (2004)

\bibitem{labelProp}
Zhu, X., Ghahramani, Z.: Learning from labeled and unlabeled data with label
  propagation. Tech. rep. (2002)

\bibitem{zhu2009introduction}
Zhu, X., Goldberg, A.B.: Introduction to semi-supervised learning. Synthesis
  lectures on artificial intelligence and machine learning  \textbf{3}(1),
  1--130 (2009)

\bibitem{zomorodian2012topological}
Zomorodian, A.: Topological data analysis. Advances in applied and
  computational topology  \textbf{70},  1--39 (2012)

\bibitem{computingPersistHomology}
Zomorodian, A., Carlsson, G.: Computing persistent homology. Discrete \&
  Computational Geometry  \textbf{33}(2),  249--274 (2005)

\end{thebibliography}

%%%%%%%%%%%%%%%%%%%%%%%%%%%%%%%%%%%%%%%%%%%%%%%%%5
\appendix
\section{Tables}\label{app:tables}

\begin{table}[ht]
\centering
\resizebox{\linewidth}{!}{%
{\small

\begin{tabular}{c|cc|cc|cc}
 \toprule
  & \multicolumn{2}{c|}{Blobs} & \multicolumn{2}{c|}{Circles} & \multicolumn{2}{c}{Moons}\\
  Method & SVM & RF & SVM & RF & SVM & RF\\
 \midrule
 \rowcolors{0}{white}{black!10!white}
 Label Propagation & 100 & 100 & 60.0 & 100 & 84.0 & 92.0\\
 Label Spreading & 100 & 100 & 64.0 & 98.0 & 94.0 & 96.0\\
 Self Training classifier &  100 & 100 & 64.0 & 100 & 88.0 & 92.0\\
 \midrule
 Bottleneck & 100 & 100 & 60.0 & 100 & 84.0 & 100\\
 Bottleneck threshold 0.8 &  100 & 100 & 64.0 & 96.0 & 92.0 & 100\\
 Bottleneck UMAP &  100 & 100 & 64.0 & 98.0 & 86.0 & 98.0\\
 Bottleneck UMAP threshold & 100 & 100 & 62.0 & 100 & 96.0 & 100\\
 Wasserstein & 100 & 100 & 62.0 & 98.0 & 88.0 & 100\\
 Wasserstein threshold 0.8 & 100 & 100 & 74.0 & 94.0 & 88.0 & 100\\
 Wasserstein UMAP &  100 & 100 & 60.0 & 100 & 84.0 & 96.0\\
 Wasserstein UMAP threshold 0.8 &  100 & 100 & 58.0 & 100 & 86.0 & 98.0\\
 \midrule
 Connectivity1 &  100 & 100 & 50.0 & 88.0 & 92.0 & 96.0\\
 Connectivity1 UMAP & 100 & 100 & 62.0 & 100 & 96.0 & 96.0\\
 Connectivity2 &  100 & 100 & 50.0 & 94.0 & 90.0 & 96.0\\
 Connectivity2 UMAP &  100 & 100 & 66.0 & 100 & 82.0 & 96.0\\
 \bottomrule
\end{tabular}}}
\caption{Accuracy results for the SVM and RF classifiers trained with data annotated by each of the annotation methods (classical, homological and connectivity) tin the 3 synthetic datasets.}\label{tab:synthetics}
\end{table}

\begin{table}[ht]
\centering
\resizebox{\linewidth}{!}{%
{\small

\begin{tabular}{ccccc}
 \toprule
  Method & \% Correct Labelled & \% Labelled  & Accuracy SVM & Accuracy RF\\
 \midrule
 \rowcolors{0}{white}{black!10!white}
 Label Propagation & 100 & 100 & 100 & 100 \\
 Label Spreading & 100 & 100 & 100 & 100 \\
 Self Training classifier & 100 & 100 & 100 & 100 \\
 \midrule
 Bottleneck & 100 & 100 & 100 & 100 \\
 Bottleneck threshold & 100 & 100 & 100 & 100 \\
 Bottleneck UMAP & 100 & 100 & 100 & 100 \\
 Bottleneck UMAP threshold & 100 & 100 & 100 & 100 \\
 Wasserstein & 100 & 100 & 100 & 100 \\
 Wasserstein threshold & 100 & 98.5 & 100 & 100 \\
 Wasserstein UMAP & 100 & 100 & 100 & 100 \\
 Wasserstein UMAP threshold & 100 & 100 & 100 & 100 \\
 \midrule
 Connectivity1 & 100 & 91.5 & 100 & 100 \\
 Connectivity1 UMAP & 100 & 86.5 & 100 & 100 \\
 Connectivity2 & 100 & 100 & 100 & 100 \\
 Connectivity2 UMAP & 100 & 100 & 100 & 100 \\
 \bottomrule
\end{tabular}}}
\caption{Results for the Blobs dataset.}\label{tab:blobs}
\end{table}

\begin{table}[ht]
\centering
\resizebox{\linewidth}{!}{%
{\small

\begin{tabular}{ccccc}
 \toprule
  Method & \% Correct Labelled & \% Labelled  & Accuracy SVM & Accuracy RF\\
 \midrule
 \rowcolors{0}{white}{black!10!white}
 Label Propagation & 98.5 & 100 & 60.0 & 100 \\
 Label Spreading & 97.0 & 100 & 64.0 & 98.0 \\
 Self Training classifier & 100 & 100 & 64.0 & 100 \\
 \midrule
 Bottleneck & 100 & 100 & 60.0 & 100 \\
 Bottleneck threshold & 100 & 81.5 & 64.0 & 96.0 \\
 Bottleneck UMAP & 100 & 100 & 64.0 & 98.0 \\
 Bottleneck UMAP threshold & 100 & 100 & 62.0 & 100 \\
 Wasserstein & 96.5 & 100 & 62.0 & 98.0 \\
 Wasserstein threshold & 100 & 68.0 & 74.0 & 94.0 \\
 Wasserstein UMAP & 100 & 100 & 60.0 & 100 \\
 Wasserstein UMAP threshold & 100 & 100 & 58.0 & 100 \\
 \midrule
 Connectivity1 & 91.8 & 49.0 & 50.0 & 88.0 \\
 Connectivity1 UMAP & 100 & 96.0 & 62.0 & 100 \\
 Connectivity2 & 97.0 & 51.5 & 50.0 & 94.0 \\
 Connectivity2 UMAP & 100 & 100 & 66.0 & 100 \\
 \bottomrule
\end{tabular}}}
\caption{Results for the Circles dataset.}\label{tab:circles}
\end{table}

\begin{table}[ht]
\centering
\resizebox{\linewidth}{!}{%
{\small

\begin{tabular}{ccccc}
 \toprule
  Method & \% Correct Labelled & \% Labelled  & Accuracy SVM & Accuracy RF\\
 \midrule
 \rowcolors{0}{white}{black!10!white}
 Label Propagation & 95.5 & 100 & 84.0 & 92.0 \\
 Label Spreading & 99.0 & 100 & 94.0 & 96.0 \\
 Self Training classifier & 91.5 & 100 & 88.0 & 92.0 \\
 \midrule
 Bottleneck & 100 & 100 & 84.0 & 100 \\
 Bottleneck threshold & 100 & 76.5 & 92.0 & 100 \\
 Bottleneck UMAP & 100 & 100 & 86.0 & 98.0 \\
 Bottleneck UMAP threshold & 100 & 100 & 96.0 & 100 \\
 Wasserstein & 98.5 & 100 & 88.0 & 100 \\
 Wasserstein threshold & 100 & 53.5 & 88.0 & 100 \\
 Wasserstein UMAP & 100 & 100 & 84.0 & 96.0 \\
 Wasserstein UMAP threshold & 100 & 99.5 & 86.0 & 98.0 \\
 \midrule
 Connectivity1 & 94.1 & 51.0 & 92.0 & 96.0 \\
 Connectivity1 UMAP & 100 & 71.0 & 96.0 & 96.0 \\
 Connectivity2 & 93.4 & 60.5 & 90.0 & 96.0 \\
 Connectivity2 UMAP & 100 & 100 & 82.0 & 96.0 \\
 \bottomrule
\end{tabular}}}
\caption{Results for the Moons dataset.}\label{tab:moons}
\end{table}
%%%%%%%%%%%%%%%%%%%%%%%%%%%%%%%%%%%%%%%%%%%%%%%%%%%%%%%%%%%%%%%%%%%%%%%%%

\begin{table*}[ht]
\centering
\resizebox{\linewidth}{!}{%
{\small

\begin{tabular}{c|cccc}
 \toprule
  Method & \% Correct Labelled & \% Labelled  & Accuracy SVM & Accuracy RF\\
 \midrule
 \rowcolors{0}{white}{black!10!white}
 Base & - & - & 71.6 & 70.2 \\
 \midrule
 Label Propagation & 63.5 & 100 & 62.7 & 62.7 \\
 Label Spreading & 63.5 & 100 & 62.7 & 62.7 \\
 Self Training classifier & 64.9 & 100 & 52.2 & 52.2 \\
 \midrule
 Bottleneck & 41.9 & 100 & 47.8 & 47.8 \\
 Bottleneck threshold 0.8 & 0 & 0 & 70.2 & 64.2 \\
 Bottleneck threshold 0.6 & 0 & 0 & 70.2 & 64.2 \\
 Bottleneck threshold 0.4 & 41.9 & 100 & 47.8 & 47.8 \\
 Bottleneck threshold 0.2 & 41.9 & 100 & 47.8 & 47.8 \\
 Bottleneck UMAP & 81.8 & 100 & 71.6 & 68.7 \\
 Bottleneck UMAP threshold 0.8 & 85.9 & 48.0 & 70.2 & 64.2\\
 Bottleneck UMAP threshold 0.6 & 85.3 & 78.4 & 74.6 & 79.1\\
 Bottleneck UMAP threshold 0.4 & 81.7 & 100 & 71.6 & 68.7\\
 Bottleneck UMAP threshold 0.2 & 81.7 & 100 & 71.6 & 68.7\\
 Wasserstein & 77.0 & 100 & 73.1 & 77.6 \\
 Wasserstein threshold 0.8 & 0 & 0 & 70.2 & 64.2 \\
 Wasserstein threshold 0.6 & 100 & 6.8 & 71.6 & 68.7\\
 Wasserstein threshold 0.4 & 77.0 & 100 & 73.1 & 77.6\\
 Wasserstein threshold 0.2 & 77.0 & 100 & 73.1 & 77.6 \\
 Wasserstein UMAP & 81.1 & 100 & 74.6 & 73.1\\
 Wasserstein UMAP threshold 0.8 & 89.4 & 44.6 & 68.7 & 59.7 \\
 Wasserstein UMAP threshold 0.6 & 83.6 & 82.4 & 73.1 & 65.7 \\
 Wasserstein UMAP threshold 0.4 & 81.1 & 100 & 74.6 & 73.1 \\
 Wasserstein UMAP threshold 0.2 & 81.1 & 100 & 74.6 & 73.1 \\
 \midrule
 Connectivity1 & 96.4 & 37.8 & 70.2 & 61.2 \\
 Connectivity1 UMAP & 93.7 & 42.6 & 71.6 & 59.7 \\
 Connectivity2 & 96.8 & 42.6 & 70.2 & 59.7 \\
 Connectivity2 UMAP & 92.8 & 46.6 & 70.2 & 64.2 \\
 \bottomrule
\end{tabular}}}
\caption{Results for the LiverGenderAL dataset.}\label{tab:liveral}
\end{table*}

\begin{table*}[ht]
\centering
\resizebox{\linewidth}{!}{%
{\small

\begin{tabular}{c|cccc}
 \toprule
  Method & \% Correct Labelled & \% Labelled  & Accuracy SVM & Accuracy RF\\
 \midrule
 \rowcolors{0}{white}{black!10!white}
 Base & - & - & 80.3 & 78.9 \\
 \midrule
 Label Propagation & 82.5 & 100 & 82.9 \\
 Label Spreading & 82.5 & 100 & 82.9 \\
 Self Training classifier & 69.5 & 100 & 59.2 & 57.9 \\
 \midrule
 Bottleneck & 49.2 & 100 & 53.9 & 53.9\\
 Bottleneck threshold 0.8 & 0 & 0 & 80.3 & 80.3 \\
 Bottleneck threshold 0.6 & 0 & 0 & 80.3 & 80.3 \\
 Bottleneck threshold 0.4 & 49.2 & 100 & 54.0 & 54.0 \\
 Bottleneck threshold 0.2 & 49.2 & 100 & 54.0 & 54.0 \\
 Bottleneck UMAP & 75.7 & 100 & 80.3 & 80.3 \\
 Bottleneck UMAP threshold 0.8 & 94.6 & 52.5 & 81.6 & 80.3\\
 Bottleneck UMAP threshold 0.6 & 86.3 & 70.1 & 81.6 & 80.3\\
 Bottleneck UMAP threshold 0.4 & 75.7 & 100 & 80.3 & 80.3\\
 Bottleneck UMAP threshold 0.2 & 75.7 & 100 & 80.3 & 80.3\\
 Wasserstein & 81.9 & 100 & 82.9 & 82.9 \\
 Wasserstein threshold 0.8 & 0 & 0 & 80.3 & 78.9\\
 Wasserstein threshold 0.6 & 100 & 45.8 & 82.9 & 82.9\\
 Wasserstein threshold 0.4 & 81.9 & 100 & 82.9 & 82.9\\
 Wasserstein threshold 0.2 & 81.9 & 100 & 82.9 & 82.9 \\
 Wasserstein UMAP & 72.9 & 100 & 65.8 & 86.8\\
 Wasserstein UMAP threshold 0.8 & 94.4 & 50.9 & 82.9 & 76.3 \\
 Wasserstein UMAP threshold 0.6 & 79.9 & 75.7 & 81.6 & 79.0 \\
 Wasserstein UMAP threshold 0.4 & 72.9 & 100 & 65.8 & 86.8 \\
 Wasserstein UMAP threshold 0.2 & 72.9 & 100 & 65.8 & 86.8 \\
 \midrule
 Connectivity1 & 80.2 & 57.1 & 46.1 & 46.1 \\
 Connectivity1 UMAP & 93.7 & 35.6 & 75.0 & 73.7 \\
 Connectivity2 & 81.1 & 59.9 & 46.1 & 46.1 \\
 Connectivity2 UMAP & 90.9 & 43.5 & 76.3 & 72.4 \\
 \bottomrule
\end{tabular}}}
\caption{Results for the LiverGenderCR dataset.}\label{tab:livercr}
\end{table*}

\begin{table*}[ht]
\centering
\resizebox{\linewidth}{!}{%
{\small

\begin{tabular}{c|cccc}
 \toprule
  Method & \% Correct Labelled & \% Labelled  & Accuracy SVM & Accuracy RF\\
 \midrule
 \rowcolors{0}{white}{black!10!white}
 Base & - & - & 97.0 & 88.6 \\
 \midrule
 Label Propagation & 93.7 & 100 & 97.4 & 93.2 \\
 Label Spreading & 93.7 & 100 & 97.4 & 93.2 \\
 Self Training classifier & 95.7 & 100 & 95.1 & 93.6 \\
 \midrule
 Bottleneck & 90.3 & 100 & 97.4 & 90.5 \\
 Bottleneck threshold 0.8 & 100 & 15.1 & 99.2 & 92.4\\
 Bottleneck threshold 0.6 & 92.7 & 84.6 & 99.2 & 91.3\\
 Bottleneck threshold 0.4 & 90.3 & 100 & 97.4 & 90.5 \\
 Bottleneck threshold 0.2 & 90.3 & 100 & 97.4 & 90.5 \\
 Bottleneck UMAP & 98.1 & 100 & 97.4 & 96.2 \\
 Bottleneck UMAP threshold 0.8 & 100 & 83.1 & 97.4 & 94.3\\
 Bottleneck UMAP threshold 0.6 & 97.3 & 93.3 & 97.4 & 94.6\\
 Bottleneck UMAP threshold 0.4 & 98.1 & 100 & 97.4 & 96.2\\
 Bottleneck UMAP threshold 0.2 & 98.1 & 100 & 97.4 & 96.2\\
 Wasserstein & 95.0 & 100 & 97.0 & 96.2 \\
 Wasserstein threshold 0.8 & 100 & 17.5 & 97.4 & 89.8\\
 Wasserstein threshold 0.6 & 99.1 & 73.5 & 99.2 & 93.6\\
 Wasserstein threshold 0.4 & 95.0 & 100 & 97.0 & 96.2\\
 Wasserstein threshold 0.2 & 95.0 & 100 & 97.0 & 96.2 \\
 Wasserstein UMAP & 98.0 & 100 & 97.0 & 95.8\\
 Wasserstein UMAP threshold 0.8 & 99.2 & 78.8 & 97.4 & 95.1 \\
 Wasserstein UMAP threshold 0.6 & 99.3 & 97.7 & 97.0 & 95.5 \\
 Wasserstein UMAP threshold 0.4 & 98.0 & 100 & 97.0 & 95.8\\
 Wasserstein UMAP threshold 0.2 & 98.0 & 100 & 97.0 & 95.8 \\
 \midrule
 Connectivity1 & 94.1 & 22.4 & 93.6 & 87.9 \\
 Connectivity1 UMAP & 67.1 & 16.4 & 80.3 & 85.2 \\
 Connectivity2 & 94.7 & 25.0 & 93.6 & 87.5 \\
 Connectivity2 UMAP & 34.8 & 11.2 & 53.8 & 75.0 \\
 \bottomrule
\end{tabular}}}
\caption{Results for the Banknote dataset.}\label{tab:banknote}
\end{table*}

\end{document}